\documentclass[conference,compsoc]{IEEEtran}
\ifCLASSOPTIONcompsoc
  \usepackage[nocompress]{cite}
\else
  \usepackage{cite}
\fi
\ifCLASSINFOpdf
\else
\fi
\usepackage{graphicx}
\usepackage{booktabs}
\usepackage{amsmath}
\usepackage{xcolor}
\usepackage{mdframed}
\usepackage{amssymb}
\usepackage{flushend} 
\usepackage{url}
\usepackage{xurl}
\usepackage{listings}
\usepackage{mdframed}

\begin{document}

\title{%
  \vspace{-1.5cm} 
  {\normalfont\normalsize\centering
  2024 IEEE International Workshop on Large Language Models for Finance, PREPRINT COPY\\
  \vspace{0.5cm} 
  }
FANAL - Financial Activity News Alerting Language Modeling Framework
}


%
\author{
\IEEEauthorblockN{Urjitkumar Patel\textsuperscript{*}}
\IEEEauthorblockA{urjitkumar.patel@spglobal.com}
\IEEEauthorblockN{Fang-Chun Yeh\textsuperscript{*}}
\IEEEauthorblockA{jessie.yeh@spglobal.com}
\IEEEauthorblockN{Chinmay Gondhalekar\textsuperscript{*}}
\IEEEauthorblockA{chinmay.gondhalekar@spglobal.com}
\IEEEauthorblockN{Hari Nalluri\textsuperscript{$\dagger$}}
\IEEEauthorblockA{hari.nalluri@spglobal.com}
}
\maketitle





\begin{abstract}

In the rapidly evolving financial sector, the accurate and timely interpretation of market news is essential for stakeholders needing to navigate unpredictable events. This paper introduces FANAL (Financial Activity News Alerting Language Modeling Framework), a specialized BERT-based framework engineered for real-time financial event detection and analysis, categorizing news into twelve distinct financial categories. FANAL leverages silver-labeled data processed through XGBoost and employs advanced fine-tuning techniques, alongside ORBERT (Odds Ratio BERT), a novel variant of BERT fine-tuned with ORPO (Odds Ratio Preference Optimization) for superior class-wise probability calibration and alignment with financial event relevance. We evaluate FANAL's performance against leading large language models, including GPT-4o, Llama-3.1 8B, and Phi-3, demonstrating its superior accuracy and cost efficiency. This framework sets a new standard for financial intelligence and responsiveness, significantly outstripping existing models in both performance and affordability.

\end{abstract}


\noindent{\footnotesize\textsuperscript{*}Authors contributed equally to this research.\\
\textsuperscript{$\dagger$}Author provided critical revisions and domain insights to enhance the manuscript.}




\section{Introduction}

In today's financial markets, the vast nature of financial activities generates a constant stream of news, making it challenging to stay updated. The volume of data can overwhelm traditional methods and impact strategic decision-making.

Timely and accurate categorization of financial news is crucial for delivering real-time information that influences investor decisions and market dynamics. Our framework categorizes financial news into twelve targeted categories, enhancing market readiness and strategic decision-making.

Advancements in Natural Language Processing (NLP), particularly through transformer-based models like BERT \cite{devlin2019bert}, T5 \cite{raffel2023exploring}, GPT \cite{openai2023gpt4}, and Llama-3.1 \cite{llama3modelcard}, have significantly enhanced our ability to process complex datasets. Despite their capabilities, the high computational demands, infrastructure costs, and API service fees associated with deploying and using such models often make them impractical for many applications.

Motivated by the achievements of CANAL (Cyber Activity News Alerting Language Model) \cite{Patel_2024}, we present FANAL (Finance Activity News Alerting Language Modeling Framework), employing sophisticated fine-tuning methods like ORPO (Odds Ratio Preference Optimization) \cite{hong2024orpo} and specialized financial news datasets. In this research, we introduce ORBERT, a BERT-based model fine-tuned with ORPO to improve the handling of class imbalances and refine predictions according to desired class distributions, justifying the appellation ORBERT. FANAL offers a cost-effective, robust solution, delivering enhanced categorization performance with limited training data.
Key contributions of our research include:

\begin{itemize}
    \item \textbf{Introduction of FANAL:} A state-of-the-art framework designed specifically for efficient and accurate categorization of financial news.
    \item \textbf{Advanced Categorization Scheme:} FANAL segments financial news into twelve precise categories, each tailored to a specific type of financial activity, enhancing the granularity of data analysis.
    \item \textbf{Introduction of ORBERT:} Streamlined fine-tuning of BERT with ORPO (Odds Ratio Preference Optimization) to achieve balanced performance across all classes and enhance the model's ability to distinguish between closely related categories.
    \item \textbf{Benchmarking Against Larger Models:} We provide a comparative analysis with major LLMs such as GPT-4o \& Llama-3.1, highlighting FANAL's efficiency in both zero-shot and few-shot LLM settings.
    \item \textbf{Cost-Effective and Resource-Efficient Solution:} Demonstrating FANAL as a more economical and sustainable option in financial news analysis.
    \item \textbf{Real-Time Financial News Utilization:} Leveraging live data feeds to maintain up-to-date and comprehensive market insights.
\end{itemize}

\section{Literature Review}

The finance sector is undergoing a transformation due to escalating complexities in financial activities. Cao \cite{cao2021ai} outlines the evolution of AI from traditional analytics to advanced data science, emphasizing its growing importance in effective financial monitoring and management. This dynamic landscape necessitates robust AI-driven tools to enhance decision-making and risk management.

The advent of transformer-based models like BERT \cite{devlin2019bert} and its financial derivative, FinBERT \cite{araci2019finbert}, has significantly advanced NLP, enhancing tasks like sentiment analysis. The proliferation of generative models such as GPT \cite{openai2023gpt4}, Gemini \cite{team2023gemini}, BloombergGPT \cite{wu2023bloomberggpt}, Claude \cite{anthropic2024claude}, Llama \cite{llama3modelcard}, Mixtral \cite{jiang2023mistral} and others, has reshaped industries, despite the high computational demands limiting some applications in finance \cite{Ganesh_2021, floridi2020gpt3}. The industry is adapting by developing efficient models like Phi-3 \cite{abdin2024phi3}, which balance performance with resource demands.

Advancements in quantization and parameter-efficient fine-tuning methods like PEFT \cite{liu2022fewshot} and LoRA \cite{hu2021lora} are critical in adapting larger models to constrained environments. Techniques like RoSA \cite{nikdan2024rosa}, ORPO \cite{hong2024orpo}, KTO (Knowledge Transfer Optimization) \cite{ethayarajh2024ktomodelalignmentprospect}, SimPO (Simple Preference Optimization) \cite{meng2024simposimplepreferenceoptimization}, and CTO (Contrastive Preference Optimization) \cite{xu2024contrastivepreferenceoptimizationpushing} further enhance model training efficiency and performance, broadening the adoption of AI in finance where prompt decision-making is essential.

The integration of NLP with financial data analytics has led to several key innovations. The F-HMTC model \cite{ijcai2020p619} utilizes a neural hierarchical multi-label text classification system to effectively categorize financial events. The unified model by Li and Zhang \cite{ijcai2020p644} enhances financial event detection and summarization accuracy through a multi-task learning strategy with a pretrained BERT and Transformer framework. Hierarchical clustering algorithms\cite{carta2021event} merge big data and NLP to provide deeper financial market insights from diverse sources, including news and social media. Advances in financial risk prediction include Ravula’s distress dictionary \cite{ravula2021bankruptcy} for bankruptcy prediction, and Zhou et al.’s event-driven trading strategy \cite{zhou2021trade}, which utilizes NLP to predict stock movements from news articles, illustrating NLP’s practical use in real-time market strategies.

Despite these developments, challenges remain in achieving the granularity required for precise decision-making and adapting the latest fine-tuning methods. There is also a lack of comprehensive benchmarks comparing these specialized models to more generalized, resource-intensive LLMs.

The introduction of CANAL \cite{Patel_2024} represented a significant advancement, establishing a minimalist yet highly effective BERT framework for cyber activity detection with reduced computational demands. Built on a similar foundation, FANAL enhances financial news analysis by utilizing finance-specific data to identify crucial information through detailed event categorization, all within a validated, cutting-edge framework.

\begin{table*}[t]
\centering
\scriptsize
\renewcommand{\arraystretch}{1.3}
\caption{Details of the Datasets Used}
\label{tab:Details of the Datasets Used}
  \begin{tabular}{llllrr}
    \toprule
    \textbf{No} & \textbf{Dataset Name} & \textbf{Sourced} & 
    \textbf{Format} & \textbf{Period} & \textbf{Rows} \\
    \midrule
    \texttt 1 & Google News Feed & Diverse websites & Title, Snippet  &    2023-10-01 to 2024-04-30 & ~350k\\
    \midrule
    \texttt 2 & Hugging Face     & PR Newswire and  &                 &                             &       \\
    & nickmuchi/trade-the-event-finance & Business Wire & Title, Body &    2020-01-01 to 2021-12-31 &  ~304k\\
    \midrule
    \texttt 3 & Hugging Face     &                  & &                             &       \\
    & ugursa/Yahoo-Finance-News-Sentences & Yahoo Financial News & Snippet &2023-06-12 to 2023-12-20 & ~25k\\
    \bottomrule
  \end{tabular}
\end{table*}

\section{Background And Theory}

\subsection{Problem Statement}

In finance, managing the influx of thousands of daily news articles is a substantial challenge. This study addresses it by categorizing articles into twelve targeted categories, offering an efficient and cost-effective alternative to large language models through advanced computational techniques.

The categorization is as follows:
\begin{itemize}
\item \textbf{M\&A}: The process of combining two or more companies through various types of financial transactions, such as mergers, acquisitions, consolidations, or takeovers.
\item \textbf{Public Market Finance}: Refers to both borrowing money that must be repaid over time and the raising of capital by companies through the sale of securities, such as stocks or bonds, to the public on stock exchanges or other public markets.
\item \textbf{Private Placement}: The sale of stocks, bonds, or securities directly to a private investor, rather than as part of a public offering.
\item \textbf{IPO}: Initial Public Offering; the process through which a privately-held company offers its shares to the public for the first time, allowing it to raise capital from public investors.
\item \textbf{Strategic Alliances}: Collaborative agreements between independent entities aimed at achieving mutually beneficial objectives through shared resources and capabilities.
\item \textbf{Company Reorganization and Structure Change}: The process of modifying a company's organizational setup and operational framework to adapt to market dynamics or achieve strategic goals.
\item \textbf{Spin-Off/Split-Off}: The creation of a new, independent company through the sale or distribution of shares of an existing business division or subsidiary to shareholders.
\item \textbf{Dividend}: A payment made by a corporation to its shareholders, usually in the form of cash or additional shares, representing a portion of the company's profits.
\item \textbf{Credit Rating}: An assessment of the creditworthiness of a borrower, typically issued by credit rating agencies, indicating the likelihood that the borrower will repay its debt obligations in a timely manner.
\item \textbf{Debt Default}: Occurs when a borrower fails to meet its contractual obligations to repay its debt, such as failing to make interest or principal payments when due.
\item \textbf{Bankruptcy}: A legal process through which individuals or businesses that cannot repay their debts seek relief from some or all of their debts, usually through liquidation of assets or reorganization of debts under court supervision.
\item \textbf{Other}: Refers to a variety of financial events or instruments not covered by the above categories, such as launching new products, additions to an index, and educational content.
\end{itemize}

\subsection{Data}

To enhance our model's adaptability, we utilized three distinct data sources. From Hugging Face, we accessed the nickmuchi/trade-the-event-finance \cite{nickmuchi/trade-the-event-finance} dataset, containing over 304,000 financial articles mainly from PR Newswire and Business Wire (2020-2021), and the ugursa/Yahoo-Finance-News-Sentences \cite{ugursa/Yahoo-Finance-News-Sentences} dataset, with about 25,000 Yahoo financial sentences from June to December 2023. Additionally, we configured Google News RSS Feeds for financial terms like 'merger', 'financing', and 'IPO', leading to a collection of approximately 350,000 articles from October 2023 to April 2024, archived with full metadata. Dataset specifics are detailed in Table~\ref{tab:Details of the Datasets Used}.

\subsection{Theoretical Framework}

Our research focuses on categorizing news content into one of twelve financial categories, representing each financial news sentence as a sequence of \( N \) tokens \( w_1, w_2, \ldots, w_N \). We model a probability distribution \( \mathbf{P} \) over these categories based on the input tokens:

\[
\mathbf{P} = \mathbf{f}(w_1, w_2, \ldots, w_N)
\]

Where \( \mathbf{f} \) is the function estimating this distribution, ensuring:

\[
\sum_{i=1}^{12} P_i = 1, \quad \text{where} \quad P_i \text{ represents each category probability.}
\]

The final category prediction for each news item is determined by a decision function \( g \):

\[
\text{Predicted Category} = g(\mathbf{P}), \quad \text{where} \quad g: \mathbf{P} \rightarrow \text{Category}
\]

Function \( g \) either selects the highest probability or applies a threshold to ascertain the most likely category.

\section{Methodology}

\subsection{XGB Silver Labeling}

Silver labeling bridges the gap between the high accuracy of gold standard data and the scalability of unsupervised predictions, providing a cost-effective labeling method. It broadens the training dataset beyond manual annotations.

We employed XGBoost \cite{Chen_2016}, known for its performance and efficiency with smaller datasets. XGBoost optimizes this objective function:

\[
Obj = \sum_{i=1}^{n} l(y_i, \hat{y}_i) + \sum_{k=1}^{K} \Omega(f_k)
\]

Here, \( l \) measures prediction discrepancies, and \( \Omega \) includes regularization to prevent overfitting. XGBoost's meticulous management of learning rates and column sampling allows it to effectively identify complex patterns and maintain consistency, making it effective for silver labeling in financial news.

\subsection{Generative Models}

In our study, FANAL is benchmarked against three leading Large Language Models (LLMs): GPT-4o \cite{openai2023gpt4}, Llama-3.1-8B-instruct \cite{llama3modelcard}, and Phi-3 mini \cite{abdin2024phi3}. These models were chosen for their distinct roles in the current AI landscape: GPT-4o as the best performing model, Llama-3.1 as the most advanced open source model available, and Phi-3 as the most capable and cost-effective among smaller language models (SLM).

\subsubsection{GPT-4o}
An advanced multimodal model represents the next evolution in multimodal natural language processing, excelling in both text and image integration to produce text outputs. ChatGPT-4o outperforms previous iterations and other leading language models, including Claude, Llama, and PaLM, making it a top performer in diverse applications, from rigorous academic assessments to multilingual benchmarks.

\subsubsection{Llama-3.1-8B-instruct}

A significant release from Meta AI, Llama-3.1, underscores their commitment to open-source advancements in AI technology. Llama-3.1 serves as the foundation for Llama-3.1 Instruct, a specialized variant meticulously fine-tuned to excel in instruction-following scenarios.

\subsubsection{Phi-3-mini 128k-instruct} 

This model stands out with its 3.8 billion-parameter setup, designed for high performance on par with larger models due to its training on 3.3 trillion tokens of selectively curated, high-quality data. Phi-3 Mini excels in instruction adherence and safety, making it a cost-effective yet robust AI option.

\subsection{ORPO}

In our study, we opted for Odds Ratio Preference Optimization (ORPO) over other fine-tuning methods such as Knowledge Transfer Optimization (KTO)\cite{ethayarajh2024ktomodelalignmentprospect} and Contrastive Preference Optimization (CTO)\cite{xu2024contrastivepreferenceoptimizationpushing} or Simple Preference Optimization (SimPO)\cite{meng2024simposimplepreferenceoptimization} due to its distinct advantages in handling class imbalances and specific prediction preferences. Unlike KTO, which focuses on generalization and knowledge transfer, or CPO/SimPO, which emphasize learning similarities or contrasts between inputs, ORPO directly addresses class distribution by optimizing for class odds ratios. This makes ORPO particularly suitable for tasks where class imbalances are a significant concern.

\subsubsection{Handling Class Imbalances}

One of the primary challenges in financial news categorization is managing class imbalances. Certain categories may naturally occur more frequently than others, which can lead to a bias in the model’s predictions if not addressed properly. Traditional fine-tuning methods, such as those based solely on Cross-Entropy Loss, often struggle in these scenarios because they tend to favor the more frequent classes, resulting in lower recall for less frequent classes.

\textbf{ORPO's Approach:} ORPO introduces a penalty based on the odds ratio, which inherently accounts for the relative frequency of classes. By focusing on adjusting predictions according to these odds ratios, ORPO ensures that the model does not disproportionately favor the dominant classes. Instead, it encourages a more balanced performance across all categories, particularly improving recall for underrepresented classes. Additionally, ORPO enhances the distinction between classes by optimizing the model to recognize and amplify the differences between them, reducing the likelihood of misclassification in closely related categories.

\subsection{FANAL}
In this section, we describe the BERT fine-tuning techniques employed in our study. We utilized three distinct approaches: BERT fine-tuning with Cross-Entropy Loss, BERT fine-tuning with ORPO (Odds Ratio Preference Optimization), and LoRA (Low-Rank Adaptation) PEFT (Parameter-Efficient Fine-Tuning).

\subsubsection{Fine-Tuning with Cross-Entropy Loss}

\begin{figure}
    \centering
    \scriptsize 
    \setlength{\tabcolsep}{5pt} 
    \includegraphics [width=9cm] {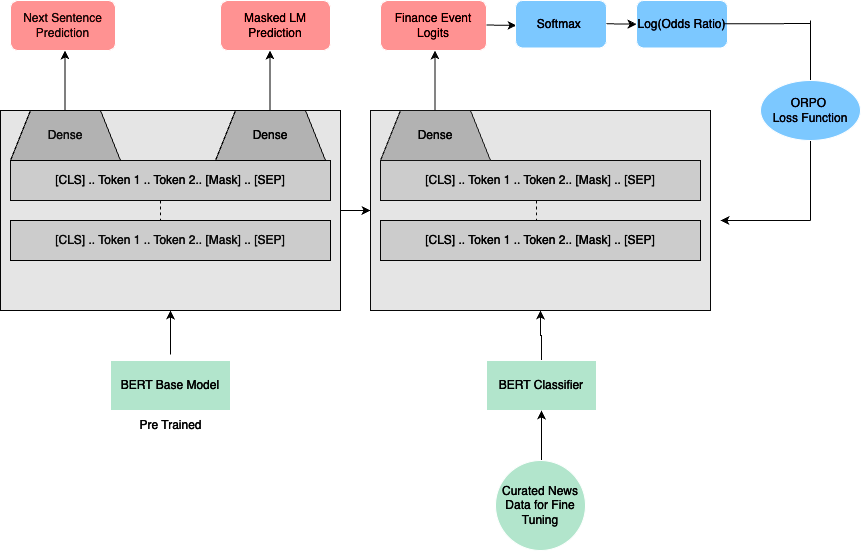}
    \caption{ORBERT Architecture}
    \label{fig:BERT}   
\end{figure}

BERT fine-tuning involves updating the parameters of a pre-trained BERT model to adapt it to a specific task. The process includes tokenization, input formatting, and optimization. The input text is tokenized and converted to indices, and special tokens like \texttt{[CLS]} and \texttt{[SEP]} are added as shown in Figure~\ref{fig:BERT}. These token indices, along with attention masks and token type ids, form the input for BERT.

The BERT model processes these inputs through multiple layers of Transformer encoders. For classification, the output corresponding to the \texttt{[CLS]} token is passed through a classification layer to predict the label using a softmax function, and the model is fine-tuned by minimizing the Cross-Entropy loss using optimization algorithms like Adam \cite{ADAM}.

\subsubsection{Fine-Tuning with ORPO Loss}

Odds Ratio Preference Optimization (ORPO) \cite{hong2024orpo} is a novel approach that enhances fine-tuning by integrating an odds ratio-based penalty with the conventional negative log-likelihood (NLL) loss. This method differentiates between favored and disfavored responses without needing a reference model. We apply ORPO to fine-tune BERT, creating a variant we call ORBERT (Odds Ratio BERT), which offers superior class-wise probability calibration.

\paragraph{\textbf{Preliminaries}}

Given an input sequence \( x \) (the sequence of tokens provided to the model), the average log-likelihood of generating an output sequence \( y \) (the predicted sequence of tokens) of length \( m \) (the number of tokens in the output sequence) is:

\begin{equation}
\log P_\theta(y|x) = \frac{1}{m} \sum_{t=1}^m \log P_\theta(y_t|x, y_{<t})
\end{equation}

where \( P_\theta(y|x) \) is the probability of generating the sequence \( y \) given the input \( x \), and \( y_t \) is the token at position \( t \) in the sequence \( y \), and \( y_{<t} \) represents the tokens before position \( t \).
\( \theta \): Model parameters (weights and biases).

The odds of generating \( y \) given \( x \) is:

\begin{equation}
\text{odds}_\theta(y|x) = \frac{P_\theta(y|x)}{1 - P_\theta(y|x)}
\end{equation}

The odds ratio between a chosen response \( y_w \) (a favored response) and a rejected response \( y_l \) (a disfavored response) is:

\begin{equation}
\text{OR}_\theta(y_w, y_l) = \frac{\text{odds}_\theta(y_w|x)}{\text{odds}_\theta(y_l|x)}
\end{equation}

\paragraph{\textbf{Objective Function of ORPO}}

The ORPO objective combines the supervised fine-tuning (SFT) loss and the relative ratio loss:

\begin{equation}
\mathcal{L}_{\text{ORPO}} = \mathbb{E}_{(x,y_w,y_l)} \left[ \mathcal{L}_{\text{SFT}} + \lambda \cdot \mathcal{L}_{\text{OR}} \right]
\end{equation}

where \( \mathcal{L}_{\text{SFT}} \) is the conventional NLL loss and \( \lambda \) is a scaling factor.

The SFT loss is the conventional NLL loss:

\begin{equation}
\mathcal{L}_{\text{SFT}} = - \log P_\theta(y|x)
\end{equation}

The relative ratio loss \( \mathcal{L}_{\text{OR}} \) maximizes the odds ratio between the favored and disfavored responses:

\begin{equation}
\label{eq:or_loss}
\mathcal{L}_{\text{OR}} = - \log \sigma \left( \log \frac{\text{odds}_\theta(y_w|x)}{\text{odds}_\theta(y_l|x)} \right)
\end{equation}

where \( \sigma \) denotes the sigmoid function.

\paragraph{\textbf{Comparison with Cross-Entropy Loss}}

The Cross-Entropy loss focuses on minimizing the difference between predicted and true labels. In contrast, ORPO not only incorporates this but also penalizes the model for generating less favored responses, ensuring a preference alignment.

\paragraph{\textbf{Gradient of ORPO}}

The gradient of the ORPO objective includes terms that penalize incorrect predictions and contrast chosen and rejected responses:

\begin{equation}
\nabla_\theta \mathcal{L}_{\text{OR}} = \delta(d) \cdot h(d)
\end{equation}

where

\begin{equation}
\delta(d) = \left[ 1 + \frac{\text{odds}_\theta(y_w|x)}{\text{odds}_\theta(y_l|x)} \right]^{-1}
\end{equation}

and

\begin{equation}
h(d) = \frac{\nabla_\theta \log P_\theta(y_w|x)}{1 - P_\theta(y_w|x)} - \frac{\nabla_\theta \log P_\theta(y_l|x)}{1 - P_\theta(y_l|x)}
\end{equation}

This approach accelerates parameter updates when the model is more likely to generate rejected responses, ensuring efficient preference alignment and reducing computational overhead while maintaining high performance.

\subsubsection{Parameter-Efficient Fine-Tuning (PEFT)}

We explore Parameter-Efficient Fine-Tuning (PEFT) \cite{liu2022fewshot} for its efficiency in fine-tuning large language models (LLMs). While full fine-tuning updates all parameters, partial fine-tuning in PEFT selectively freezes a portion of the model’s weights while fine-tuning the rest. The fine-tuning process for both full and partial parameter updates explores the performance impact on our multiclass classification task, providing insights into the trade-offs between computational efficiency and classification effectiveness.

\subsubsection{PEFT with LoRA}

We also experiment with PEFT combined with Low-Rank Adaptation (LoRA) \cite{hu2021lora}. LoRA updates a pre-trained weight matrix \( W_0 \) (the original weight matrix) with a low-rank decomposition
\begin{equation}
\label{LORA}
( W_0 + \Delta W = W_0 + BA )
\end{equation}
where \( B \in \mathbb{R}^{d \times r} \) (input dimension by rank) and \( A \in \mathbb{R}^{r \times k} \) (rank by output dimension), and the rank \( r \ll \min(d, k) \) (rank is much smaller than the input and output dimensions). \( d \) is Input dimension of the weight matrix and 
 \( k \) is the output dimension of weight matrix.
 \( r \) is Rank of the low-rank decomposition, which is much smaller than both \( d \) and \( k \). During training, \( W_0 \) is frozen, while \( A \) and \( B \) contain trainable parameters. The modified forward pass with LoRA is:

\begin{equation}
h = W_0 x + \Delta W x = W_0 x + BA x
\end{equation}

where \( x \) is the input feature vector, and \( h \) is the output feature vector.

Our approach integrates BERT’s architecture with PEFT and LoRA fine-tuning for effective multiclassification, as demonstrated in our methodology.


\subsection{Entity Relevance Module}
\label{sec:entity_relevance_module}

We incorporate the Entity Relevance Module in our experiment to enhance news processing by evaluating the contextual significance of entities in texts, going beyond traditional NER models that simply tag entities. For example, in the headline "Debt defaults soared, XYZ says," "XYZ" is recognized as a commentator rather than the main subject.

Relevance probabilities are determined using:
\begin{equation}
P(\text{Class 1 - Relevant}) = \sigma(W \cdot \Phi(\text{input}) + b)
\end{equation}
Here, \(\sigma\) is the sigmoid function, with \(W\), \(b\), and \(\Phi(\text{input})\) as the model parameters and input features. Detailed technical specifics are out of the scope of this paper.

\section{Training Scheme}

\subsection{Data Labeling and Categorization}

\begin{itemize}
\item \textbf{Gold Standard Dataset:}
A 'Gold Standard' dataset consisting of approximately 100 samples from each category across three sources (totaling ~1200 samples), mentioned in Table~\ref{tab:Details of the Datasets Used}, was meticulously labeled by domain experts. This dataset facilitated the initial training of the XGB model using an 80\%-20\% training-testing split.

\item \textbf{Silver Label Dataset using XGB Model:} 
After the initial training phase, the XGB model was applied to the remaining data from the three sources, excluding the previously used Gold Standard data. Records with high labeling confidence were selected to form the 'Silver Label' dataset. This dataset comprised approximately 16,000 high-confidence records, which played a crucial role in further fine-tuning the ORBERT model.

\item \textbf{Fine-Tuning Data for FANAL:}
The BERT models were fine-tuned using both gold and silver labeled datasets, comprising ~1,200 and ~16,000 samples respectively, providing a substantial amount of training samples for each category.

\item \textbf{Test Data for FANAL:}
To evaluate the FANAL framework and benchmark it against other large language models (LLMs), a diverse subset of approximately 1200 articles was used. These articles, sampled from all three datasets, were labeled by subject matter experts (SMEs). We ensured that all categories were adequately represented, enhancing the reliability of our evaluation.

\end{itemize}

\subsection{Training Scheme for XGB}


To counter potential bias in silver label generation, we filtered our data retrieval with precise SQL queries, ensuring a representative dataset for training. Essential hyperparameters were set as follows:
\begin{itemize}
    \item \texttt{objective='multi:softprob'}
    \item \texttt{booster='gbtree'}
    \item \texttt{lambda=1}
    \item \texttt{max\_depth=6}
\end{itemize}

\begin{figure}
\centering
\scriptsize 
\setlength{\tabcolsep}{5pt} 
\includegraphics[width=0.5\textwidth]{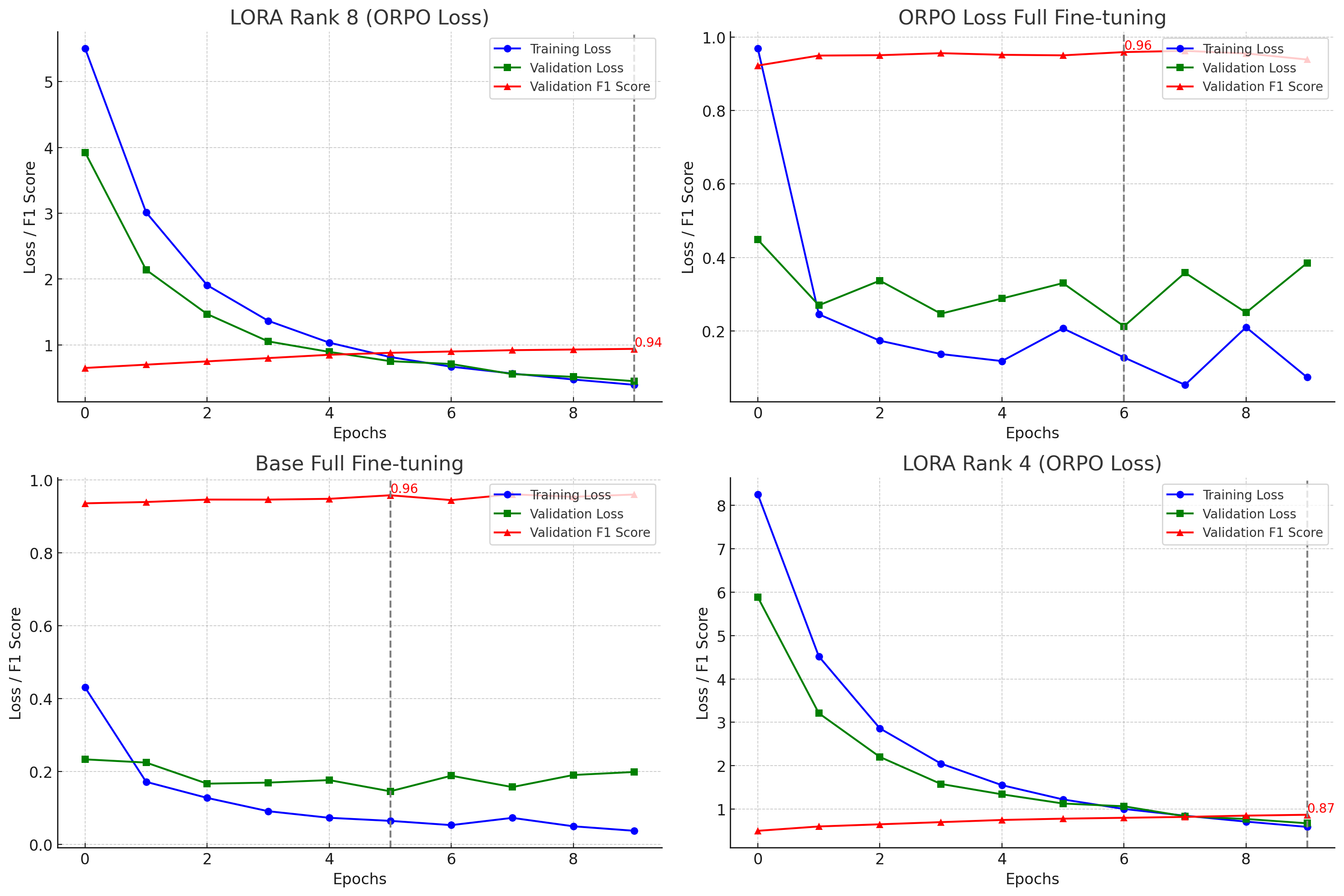}
\caption{Illustration of training and validation loss over 10 epochs}
\label{fig:loss_plots}
\end{figure}

\subsection{Training Scheme for BERT Fine-Tuning}

Three main fine-tuning runs were performed all for 10 epochs:

\begin{itemize}
\item Vanilla BERT fine-tuning: with Cross-Entropy Loss.
\item ORBERT fine-tuning: with ORPO Loss.
\item PEFT LoRA fine-tuning (Rank 4 and 8).
\end{itemize}

\paragraph{\textbf{With Cross-Entropy Loss}}

This method adjusts all layers of the BERT model to the specific task, leading to improved performance due to comprehensive tuning. However, this approach is computationally expensive and requires significant resources.

\paragraph{\textbf{With ORPO Loss}}

The ORPO (Odds Ratio Preference Optimization) loss method was applied for full BERT fine-tuning, resulting in the ORBERT model. This method optimizes the training process for robustness and performance, potentially resulting in better generalization on unseen data.

\paragraph{\textbf{PEFT LoRA Fine-Tuning}}

Parameter Efficient Fine-Tuning (PEFT) with LoRA (Low-Rank Adaptation) This approach fine-tunes only a subset of the model parameters, reducing computational costs and speeding up the training process while still achieving competitive performance.

The lowest validation losses and highest F1 scores were recorded as illustrated in Figure \ref{fig:loss_plots}:
These results indicate that different fine-tuning strategies can lead to varying rates of convergence and performance. The ORPO method and PEFT LoRA achieved their best validation loss at different epochs, while regular BERT fine-tuning reached its optimal performance earlier.
The results highlight the effectiveness of different fine-tuning strategies, with PEFT LoRA methods showing competitive performance with reduced computational costs compared to full BERT fine-tuning and ORPO Bert fine-tuning.

\subsection{ORBERT Hyperparameters}

Table \ref{tab:bert-hyperparameters} lists the key hyperparameters and their values that we used for the final ORBERT training. We employed a grid search on a small subset of data to identify the best parameters and then applied these optimal settings for the full tuning of the model.

\begin{table}
\centering
\scriptsize 
\setlength{\tabcolsep}{5pt} 
\caption{ORBERT Hyperparameters}
\label{tab:bert-hyperparameters}
\begin{tabular}{@{}ll@{}}
\toprule
\textbf{Hyperparameter}  & \textbf{Value}    \\ 
\midrule
Learning Rate            & $5e-5$            \\
Batch Size               & $32$              \\
Epochs                   & $10$              \\
Loss Function            & ORPO Loss  \\
Optimizer                & AdamW             \\
\bottomrule
\end{tabular}
\end{table}

\subsection{Prompt Engineering for LLMs}

In our approach to classifying financial news into twelve distinct categories, we systematically enhanced prompt templates across successive iterations to augment the comprehension and efficacy of Large Language Models (LLMs). This process was designed to gradually refine the model's ability to interpret and classify complex financial content accurately.

\begin{itemize}

\item \textbf{Template 1, Zero Shot: Basic Instruction Set}

Initial template aimed to gauge the foundational understanding of financial terms. It asks the model to classify news content by discerning relevant financial entities, with a default to "Other" if no specific entities are identified.

\begin{mdframed}[linewidth=0.8pt, linecolor=black, backgroundcolor=gray!8, roundcorner=8pt]
    \textbf{Template 1 - Zero Shot:}
    
    \textit{You are a financial analyst. Classify the news sentences into one of the twelve categories mentioned below and return only the category name. An entity could be an organization, a location, a place, a person, or a group. If no entities are tagged in the sentence, classify it as the "Other" category.}
    
    \vspace{2mm}
    
    \textit{Categories: M\&A, Public Market Finance, Private Placement, IPO, Strategic Alliances, Company reorganization and structure change, Dividend, Credit Rating, Spin-Off/Split-Off, Debt Default, Bankruptcy, Other.}
\end{mdframed}

\item \textbf{Template 2, Zero Shot with Definitions}

Enhances interpretive capabilities by providing precise definitions for each category, aiding the LLM in distinguishing between subtle nuances.

\vspace{3mm}

\begin{mdframed}[linewidth=0.8pt, linecolor=black, backgroundcolor=gray!8, roundcorner=8pt]
    \textbf{Template 2 - Zero Shot with Definitions:}
    
    \textit{You are a financial analyst. Classify the news sentences into one of the twelve categories mentioned below ...}
    \begin{itemize}
        \item \textit{\textbf{M\&A}: Mergers and Acquisitions - consolidation of companies or assets through various forms of financial transactions, including mergers, acquisitions, consolidations, and purchase of assets.}
        
        ...
        
        \item \textit{\textbf{Other}: For content that does not fit into the specified categories, encompassing a broad range of general financial topics not tied to specific entities.}
    \end{itemize}
\end{mdframed}

\begin{figure}[b]
    \centering
    \includegraphics [width=9cm] {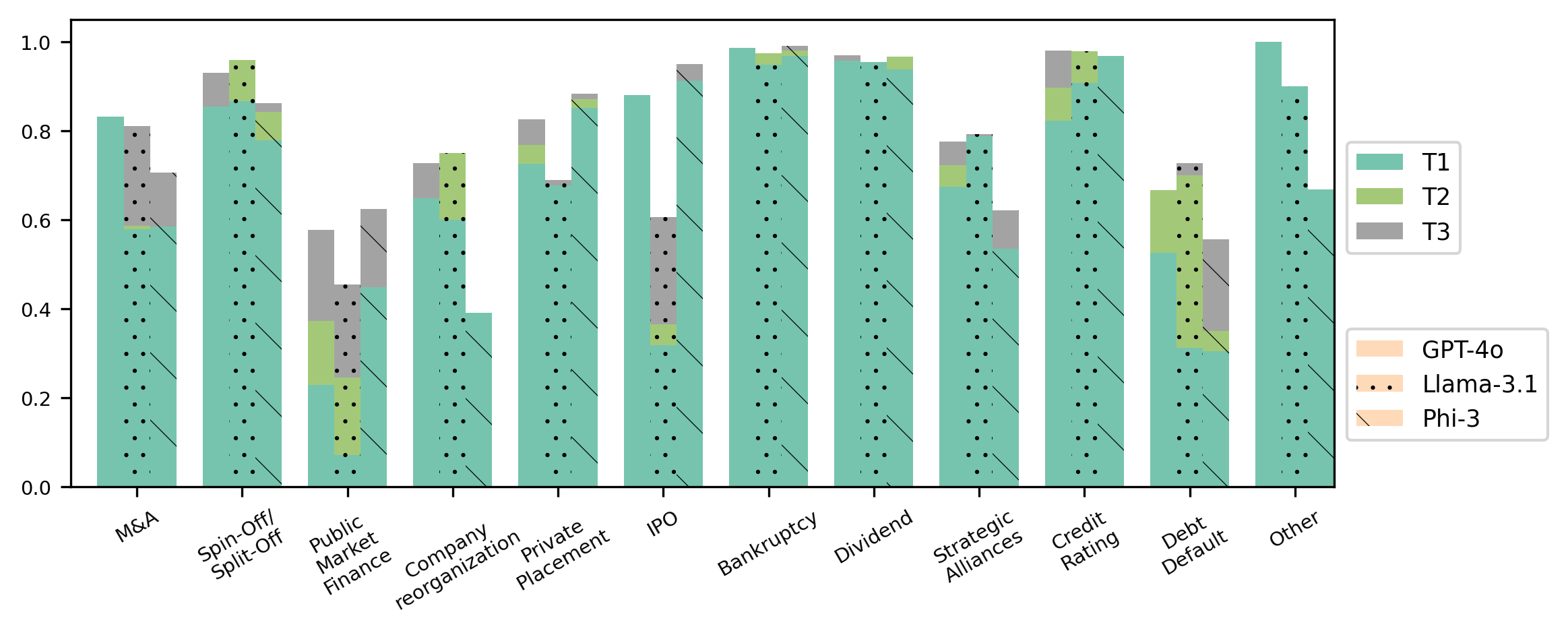}
    \caption{Precision Improves with Prompt Engineering}
    \label{fig:plot_group_stacked_bar_chart_pc}   
\end{figure}

\item \textbf{Template 3, Few Shots with Definitions and Examples}

This version builds on Template 2 by adding real examples for each category, enhancing the LLM's understanding and classification accuracy through a few-shot approach that demonstrates categorization criteria directly.

\begin{mdframed}[linewidth=0.8pt, linecolor=black, backgroundcolor=gray!8, roundcorner=8pt]
    \textbf{Template 3 - Few Shot with Definitions:}
    
    \textit{You are a financial analyst. Classify the news sentences into one of the twelve categories mentioned below ...}
    \begin{itemize}
    \item \textit{\textbf{M\&A}: Mergers and Acquisitions - consolidation of companies or assets through various...}    
    
    ...
    
    \item \textit{\textbf{Other}: For content that does not fit into the specified categories...}
    \end{itemize}

    \textbf{Examples for each category are:}
    \begin{lstlisting}[basicstyle=\ttfamily\footnotesize, breaklines=true]
    {"sentence": "Hardesty \& Hanover Acquires Corven Engineering.", "category": "M&A"},
    {"sentence": "ATHA Energy increases private placement offering up to \$22.84M.", "category": "Private Placement"},
    ...
    \end{lstlisting}
\end{mdframed}
\end{itemize}

\begin{figure}[b]
    \centering
    \includegraphics [width=9cm] {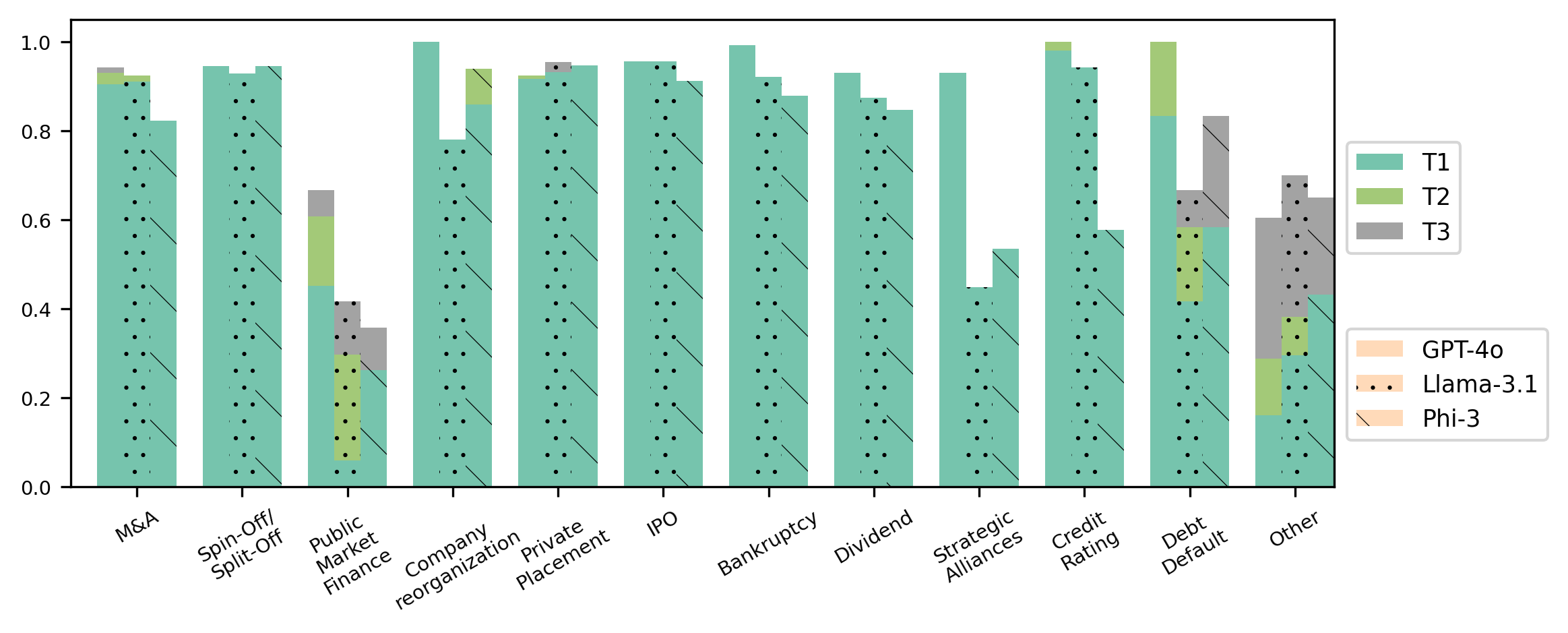}
    \caption{Recall Improves with Prompt Engineering}
    \label{fig:plot_group_stacked_bar_chart_rc}   
\end{figure}


\section{Evaluation Results}
\subsection{Evaluation Metrics}

In this section, we outline the performance metrics utilized to assess the models in our multi-class classification task. FANAL is designed to optimize \textbf{ORPO loss} Equation \ref{eq:or_loss} across 12 categories in a multi-class categorization context. For a comparative analysis with other LLMs, where probability distributions are inaccessible, we adopt standard metrics such as Precision, Recall, F1-Score, and Accuracy.

\textbf{Precision:}
\[
\text{Precision} = \frac{\text{True Positives}}{\text{True Positives} + \text{False Positives}}
\]

\textbf{Recall (Sensitivity):}
\[
\text{Recall} = \frac{\text{True Positives}}{\text{True Positives} + \text{False Negatives}}
\]

\textbf{F1-Score:}
\[
\text{F1-Score} = \frac{2 \cdot \text{Precision} \cdot \text{Recall}}{\text{Precision} + \text{Recall}}
\]

\textbf{Accuracy:}
\[
\text{Accuracy} = \frac{\text{Number of Correct Predictions}}{\text{Total Number of Predictions}}
\]

\subsection{Evaluation of XGB Model}

For all twelve financial activities, we adjust the probability thresholds of the XGBoost classifier based on performance against gold-labeled test data to ensure each category's precision meets at least a 95\% threshold. Table \ref{tab:xgb_threshold} displays the threshold values for each category.

\begin{table}
\centering
\scriptsize
\setlength{\tabcolsep}{5pt} 
\caption{XGB Probability Thresholds}
\label{tab:xgb_threshold}
\begin{tabular}{lc}
\toprule
\textbf{Category} & \textbf{Threshold} \\ 
\midrule
M\&A                  & 0.96 \\
Public Market Finance & 0.98 \\
Private Placement     & 0.80 \\
IPO                   & 0.90 \\
Spin-Off/Split-Off    & 0.88 \\
Dividend              & 0.90 \\
Credit Rating         & 0.88 \\
Debt Default          & 0.75 \\
Bankruptcy            & 0.70 \\
Other                 & 0.90 \\
Strategic Alliances   & 0.90 \\
Company Reorganization & 0.90 \\
\bottomrule
\end{tabular}
\end{table}

\begin{figure}[b]
    \centering
    \includegraphics [width=9cm] {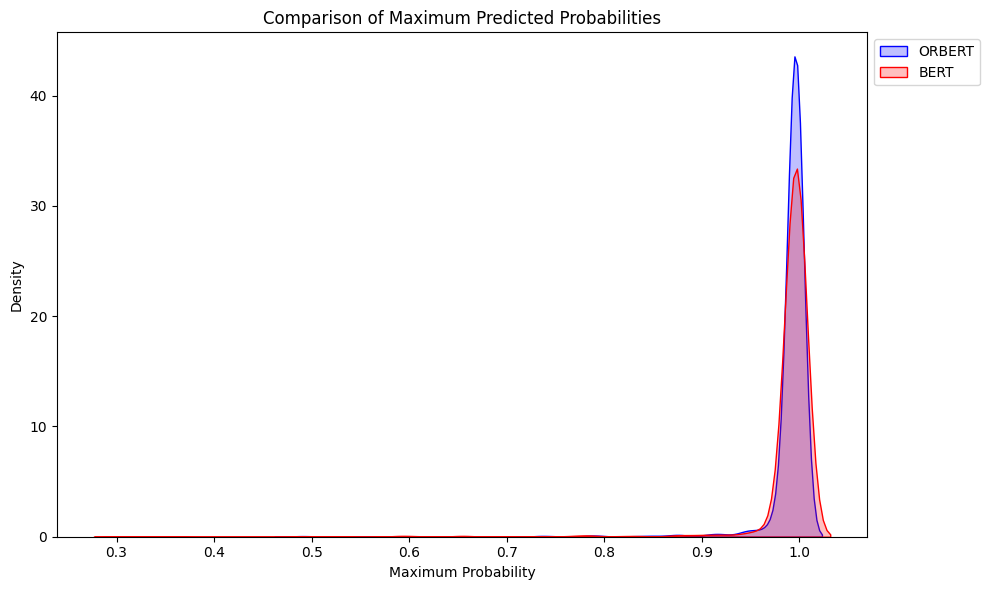}
    \caption{Fine-Tuned ORBERT vs Fine-Tuned Base BERT}
    \label{fig:orbertvbert}   
\end{figure}

\subsection{Evaluation of Expensive LLMs}



Fig.~\ref{fig:plot_group_stacked_bar_chart_pc} and Fig.~\ref{fig:plot_group_stacked_bar_chart_rc} show LLM precision and recall scores across all twelve financial categories using Template 1 (T1), Template 2 (T2), and Template 3 (T3) as detailed in section 5.5. We observed that performance improves noticeably for all evaluated LLMs from Template 1 to Template 3, from zero-shot to few-shot, especially in categories like Public Market Finance and Debt Default. Interestingly, for categories like Bankruptcy and Dividend, the LLMs already demonstrate high precision and recall in the Template 1 scenario, indicating that LLMs are proficient in understanding these categories and do not require additional support.

Moreover, GPT-4o stands out among the three LLMs but faces challenges in categories like "Other", Public Market Finance, and Debt Default. In the "Other" category, its high precision contrasts with a low recall rate, resulting in many false negatives. Conversely, in Debt Default, GPT-4o shows higher recall than precision, indicating a propensity to over-detect instances. The Public Market Finance category also proves difficult, with both low precision and recall, leading to frequent misidentification and oversight of relevant cases.

Llama-3.1 slightly underperforms compared to GPT-4o in most categories and faces similar struggles. An interesting finding about Phi-3, despite its varied performance, is its superior ability over GPT-4o in identifying Private Placement finance events in zero-shot scenarios. However, its results often deviate from the desired structure, requiring more effort to extract predictions, especially with longer prompts.

\subsection{Evaluation of BERT Fine-Tuning schemes}

Upon analyzing the loss and F1 scores in Fig.~\ref{fig:loss_plots} across all models, it is clear that full fine-tuning of BERT with ORPO loss exhibited superior performance compared to other approaches. Full fine-tuning with ORPO consistently outperformed alternatives, while fine-tuning only the \textbf{LORA (rank = 8)} layer also yielded notable results. These methods are advantageous in scenarios with constrained resources or large models, making full fine-tuning impractical.

FANAL utilizes the ORBERT model (BERT fine-tuned with ORPO loss), which shows superior ability to confidently predict categories. Fig.~\ref{fig:orbertvbert} demonstrates that ORBERT assigns higher confidence to its predictions, leading to more precise categorization, especially in closely related financial events. By leveraging ORBERT, FANAL balances performance and efficiency, ensuring robust operation in resource-constrained environments. Table~\ref{tab:orbert_vs_bert_examples} provides examples where both ORBERT and Vanilla BERT (using regular Cross-Entropy loss) correctly predicted categories, but ORBERT exhibited significantly higher confidence, reducing uncertainty and improving reliability.

\begin{table}[t]
\centering
\scriptsize
\renewcommand{\arraystretch}{1.8} 
\caption{ORBERT vs Vanilla BERT: Confidence Probabilities Comparison}
\label{tab:orbert_vs_bert_examples}
\begin{tabular}{p{6cm}cc}
\toprule
\textbf{News sample} & \textbf{ORBERT} & \textbf{BERT} \\
                  & \textbf{prob}   & \textbf{prob} \\
\midrule
\scriptsize{SiriusXM Agrees To Merge With Liberty Media Tracking Stock. Under the deal, Liberty Media will split off its ... Class - \textcolor{red}{Spin-Off/Split-Off}} 
& \textbf{0.9851} & \textbf{0.6927} \\
\midrule
\scriptsize{Capri Holdings Ltd (CPRI): The luxury fashion company's lagging sales put its 2024 merger at risk despite 14\% ... Class - \textcolor{red}{M\&A}}
& \textbf{0.9707} & \textbf{0.8270} \\
\bottomrule
\end{tabular}
\end{table}

Comparison of FANAL and other LLMs is shown in Table \ref{tab:model_performance}. This comparison highlights FANAL's ability to effectively distinguish between closely related categories.


\begin{table*}
\centering
\scriptsize
\setlength{\tabcolsep}{4pt} 
\caption{Model Performance Comparison}
\label{tab:model_performance}
\scriptsize 
\setlength{\tabcolsep}{2pt} 
\begin{tabular}{llrrrrrrrrrrrr}
\toprule
\textbf{Model} & \textbf{Matrix} & \textbf{Other} & \textbf{M\&A} & \textbf{Spin-Off/} & \textbf{Pub. Market} & \textbf{Company} &  \textbf{Private} & \textbf{IPO} & \textbf{Bankruptcy} & \textbf{Dividend} & \textbf{Strategic} & \textbf{Credit} & \textbf{Debt} \\ 
      &   &    &   & \textbf{Split-Off} & \textbf{Finance} & \textbf{reorg.} & \textbf{Placement} & & & & \textbf{Alliances} & \textbf{Rating}  & \textbf{Default} \\
      & Support  &      243 &      158 &     56  &     84  &      50 &      133 &      73 &      141 &      72 &      58 &      52 &    55 \\
\midrule
\textbf{GPT-4o} & Accuracy & \underline{57.65}\% & \underline{73.04\%} & \underline{88.33\%} & \underline{44.80}\% & \textbf{70.59\%} & 75.32\% & \underline{80.77\%} & \underline{93.71\%} & \underline{89.19\%} & \underline{71.23\%} & \textbf{94.34\%} & \underline{66.67\% }\\
FS    & Precision & \textbf{92.45\%} & 76.41\% & \textbf{92.98\%} & \underline{57.73}\% & 72.73\% & 82.64\% & 87.50\% & \textbf{98.53\%} & \textbf{97.06\%} & 77.61\% & \textbf{98.04\%} & \underline{66.67\% }\\
      & Recall   & 60.49\% & \textbf{94.30\%} & \underline{94.64\%} & \underline{66.67\%} & \textbf{96.00}\% & 89.47\% & \underline{91.30\%} & \underline{95.04\%} & \underline{91.67\%} & 89.66\% & \textbf{96.15}\% & \textbf{100.00\%} \\ 
      & F1-Score & \textbf{73.13\%} & \textbf{84.42}\% & \textbf{93.81\%} & \underline{61.88\%} & \textbf{82.76\%} & 85.92\% & \underline{89.36}\% & \underline{96.75\%} & \underline{94.29\%} & \underline{83.20\%} & \textbf{97.09\%} & \underline{80.00\%} \\ 
\midrule
\textbf{Llama-3.1} & Accuracy & 50.60\% & 72.11\% & 76.67\% & 27.78\% & 58.73\% & 66.84\% & 55.56\% & 70.14\% & 77.33\% & 35.94\% & \underline{90.74\%} & 53.33\% \\
FS    & Precision & 64.64\% & \textbf{81.07\%} & \underline{92.00\%} & 45.45\% & \underline{74.00}\% & 69.02\% & 60.61\% & \underline{97.12\%} & \underline{95.08\%} & \underline{79.31\%} & 96.08\% & 72.73\% \\
      & Recall   & \textbf{69.96\%} & \underline{86.71\%} & 82.14\% & 41.67\% & 74.00\% & \underline{95.49\%} & 86.96\% & 71.63\% & 80.56\% & 39.66\% & \underline{94.23\%} & 66.67\% \\
      & F1-Score & \underline{67.19}\% & \underline{83.79\%} & 86.79\% & 43.48\% & 74.00\% & 80.13\% & 71.43\% & 82.45\% & 87.22\% & 52.87\% & \underline{95.15\%} & 69.57\% \\
\midrule
\textbf{Phi-3} & Accuracy & 42.70\% & 57.00\% & 69.84\% & 29.41\% & 28.77\% & \underline{81.28\%} & 79.17\% & 79.34\% & 77.33\% & 31.94\% & 41.51\% & 50.00\% \\
FS    & Precision & 55.44\% & 70.66\% & 86.27\% & \textbf{62.50\%} & 31.08\% & \textbf{88.32\%} & \textbf{95.00\%} & 99.06\% & \underline{95.08\%} & 62.16\% & 95.65\% & 55.56\% \\
      & Recall   & \underline{65.02}\% & 74.68\% & 78.57\% & 35.71\% & \underline{80.00\%} & 90.98\% & 82.61\% & 74.47\% & 80.56\% & 39.66\% & 42.31\% & 83.33\% \\
      & F1-Score & 59.85\% & 72.62\% & 82.24\% & 45.45\% & 44.69\% & \underline{89.63\%} & 88.37\% & 85.02\% & 87.22\% & 48.42\% & 58.67\% & 66.67\% \\
\midrule
\textbf{FANAL}
 & Accuracy & \textbf{58.03\%} & \textbf{85.44\%} & \textbf{96.42\%} & \textbf{76.19\%} & \underline{64.00\%} & \textbf{96.24\%} & \textbf{93.15\%} & \textbf{98.58\%} & \textbf{100.00\%} & \textbf{96.55\%} & 82.70\% & \textbf{85.46\%} \\
 \textbf{with} & Precision & \underline{76.22\%} & \underline{78.94}\% & 79.41\% & 57.14\% & \textbf{88.89\%} & \underline{87.08}\% & \underline{90.67\%} & 96.53\% & 92.30\% & \textbf{82.35\%} & \underline{97.73\%} & \textbf{95.91\%} \\
\textbf{ORBERT} & Recall & 58.03\% & 85.44\% & \textbf{96.43\%} & \textbf{76.19\%} & 64.00\% & \textbf{96.24\%} & \textbf{93.15\%} & \textbf{98.58\%} & \textbf{100.00\%} & \textbf{96.55\%} & 82.69\% & \underline{85.46\%} \\
      & F1-Score & \underline{65.88\%} & 82.07\% & \underline{87.10\%} & \textbf{65.30}\% & \underline{74.42}\% & \textbf{91.43\%} & \textbf{91.89\%} & \textbf{97.54\%} & \textbf{96.00\%} & \textbf{88.89\%} & 89.58\% & \textbf{90.39\%} \\
\bottomrule
\vspace{-6pt}
\end{tabular}

{\raggedright \textit{Note.The best results in each category are highlighted in bold, while the second-best results are underlined. FS stands for Few Shots.}\par}
\end{table*}

\begin{table}[b]
\centering
\scriptsize
\setlength{\tabcolsep}{5pt} 
\caption{Comparison of LLMs for Processing 10,000 Articles }
\label{tab:running_time_comparison}
\scriptsize 
\setlength{\tabcolsep}{1.5pt} 
\begin{tabular}{lccr}
\toprule
\textbf{Model} & {\textbf{Inference Time (hr)}} & \textbf{Infrastructure} & {\textbf{Inference Cost (\$)}} \\
\midrule
GPT-4o   & 1.579 & OpenAI API   & 204.8675 \\
Llama-3.1 & 0.642 & A100 (40 GB) &   0.8363 \\
Phi-3   & 2.254 & A100 (40 GB) &   2.9362 \\
FANAL   & 0.013 & A100 (40 GB) &   0.0017 \\
\bottomrule
\end{tabular}
\end{table}

\begin{table}[b]
\centering
\scriptsize 
\setlength{\tabcolsep}{5pt} 
\small 
\setlength{\tabcolsep}{2pt} 
\caption{Example classification snippets.}
\label{tab:Example classification snippets}
\begin{tabular}{p{4.2cm} p{4.2cm}}
\toprule
\textbf{Financial News} & 
\textbf{FANAL + Entity Relevance} \\
\midrule
\scriptsize{Frontier Shuffles the Cards and Restructures the Leadership Team - CEOWORLD magazine} & \scriptsize{\textcolor{purple}{Frontier} Shuffles the Cards and \textcolor{blue}{Restructures} the Leadership Team - CEOWORLD magazine - \textcolor{brown}{\textit{Class - Company reorganization and structure change}}} \\
\midrule
\scriptsize{Star Bulk and Eagle Bulk Shipping Agree to \$2.1B Merger | SupplyChainBrain} & \scriptsize{\textcolor{purple}{Star Bulk} and \textcolor{red}{Eagle Bulk} Shipping Agree to \$2.1B \textcolor{blue}{Merger} - \textcolor{brown}{\textit{Class - M\&A}}} \\
\midrule
\scriptsize{Evergy prices upsized \$1.2B convertible notes (NASDAQ:EVRG) | Seeking Alpha} & \scriptsize{\textcolor{purple}{Evergy} prices upsized \$1.2B \textcolor{blue}{convertible notes} (NASDAQ:EVRG) | Seeking Alpha - \textcolor{brown}{\textit{Class - Public Market Finance}}} \\
\bottomrule
\end{tabular}
\end{table}

\subsection{FANAL vs Other Expensive LLMs}

Table \ref{tab:model_performance} presents a comprehensive performance comparison between FANAL utilizing ORBERT and other LLMs. FANAL stands out for its performance on categories such as Strategic Alliances, Bankruptcy, Dividend, Debt Default, Public Market Finance, and Private Placement, consistently outperforming other models by significant margins in at least three out of four metrics.

When comparing FANAL to other models, GPT-4o FS (Few Shot) emerges as the closest competitor. While FANAL consistently outperforms GPT-4o by a significant margin in most categories, in a few categories such as M\&A and Spin-Off/Split-Off where we see a close match, FANAL demonstrates higher accuracy, whereas GPT-4o FS achieves higher precision and F1 scores. GPT-4o FS attains top performance, notably in the Credit Rating category, where it excels across all metrics. However, its performance varies significantly in categories like Public Market Finance and Debt Default, likely due to overlapping characteristics in the training data and possibly due to sensitivity to noise and outliers. In most cases, FANAL maintains high F1-scores, effectively balancing precision and recall as compared to Llama-3.1 FS and Phi-3 FS indicating weaker performance in balancing precision and recall. However, Phi-3 exhibits notable precision in the IPO and Private Placement categories.

Table \ref{tab:running_time_comparison} provides an empirical comparison of estimated inference time and cost for processing 10,000 articles between FANAL and other LLMs. With an inference time of just 0.013 hours and an inference cost of \$0.0017 on A100 (40 GB) infrastructure, FANAL proves to be both effective and highly efficient. Cost estimations are based on the current OpenAI API pricing \cite{openai_pricing_2024} and the lowest available A100 usage rates, approximately \$1.2/hour \cite{databricks_pricing_2024} \cite{a100_price_2024} at the time of writing. FANAL not only processes data faster but also maintains minimal operational costs, emphasizing its viability for large-scale applications.

Overall, FANAL's superior performance is attributed to its balanced approach, ensuring high scores across most categories and empirical efficiency. This consistency suggests that FANAL is a well-tuned framework for a wide range of prediction tasks, making it a reliable choice compared to other LLMs.

\subsection{FANAL in action}

Table \ref{tab:Example classification snippets} illustrates the integration of FANAL with the Entity Relevance Module (Section \ref{sec:entity_relevance_module}), enhancing our capability to dissect and interpret news data for entity-specific financial signals. This integrated process enables us to construct an in-depth financial insights portfolio tailored to specific entities and sectors.

\section{Conclusion}

We have demonstrated that our FANAL, trained using a silver labeling approach with just 1200 manually labeled data points, excels in 12-class financial news categorization. FANAL outperforms current top Large Language Models in most categories while being more cost-efficient. Our comparison also highlights the performance of existing industry LLMs in financial categorization, suggesting that their efficacy can be improved with prompt engineering techniques. Future research may focus on enhancing FANAL's training data quality and quantity for under-performing categories and leveraging fine-tuning techniques like ORPO to fine-tune smaller LLMs such as Llama-3.1 and Phi-3 using financial-specific datasets.



\bibliographystyle{IEEEtran}

\bibliography{sample-base}

\end{document}